\documentclass{article}
\usepackage{orcidlink}

\usepackage[a4paper, total={6in, 8in}]{geometry}

\usepackage{amsmath}
\usepackage{amssymb}
\usepackage{float}

\usepackage{url}

\usepackage{graphicx} % Required for inserting images

\usepackage{xcolor}
\usepackage{amssymb}
\usepackage{adjustbox}
\usepackage{booktabs} % To thicken table lines
\usepackage{amsmath}
\usepackage{caption}
\usepackage{multirow}
\usepackage{makecell}
\usepackage{setspace}

\usepackage{dsfont}

\newcommand{\ie}{\textit{i.e.}}
\newcommand{\eg}{\textit{e.g.}}

\usepackage{bm}
\newcommand{\PC}[1]{\ignorespaces}
\newcommand{\CP}[1]{\ignorespaces}

\newtheorem{defi}{Definition}

\title{A Theoretical and Practical Framework for Evaluating Uncertainty Calibration in Object Detection}
\author{Pedro Conde$^1$, Rui L. Lopes $^2$, Cristiano Premebida$^1$}
\date{\small $^1$ Institute os Systems and Robotics, Department Of Electrical and Computer Engineering, University of Coimbra\\
\{pedro.conde,cpremebida\}@isr.uc.pt\\
$^2$ Critical Software, S.A.\\
rui.lopes@criticalsoftware.com}

\begin{document}

\maketitle

\begin{abstract}
    \noindent The proliferation of Deep Neural Networks has resulted in machine learning systems becoming increasingly more present in various real-world applications. Consequently, there is a growing demand for highly reliable models in many domains, making the problem of uncertainty calibration pivotal when considering the future of deep learning. This is especially true when considering object detection systems, that are commonly present in safety-critical applications such as autonomous driving, robotics and medical diagnosis. For this reason, this work presents a novel theoretical and practical framework to evaluate object detection systems in the context of uncertainty calibration. This encompasses a new comprehensive formulation of this concept through distinct formal definitions, and also three novel evaluation metrics derived from such theoretical foundation. The robustness of the proposed uncertainty calibration metrics is shown through a series of representative experiments.
  
\end{abstract}

\section{Introduction}

Deep Neural Networks (DNNs) have revolutionized the applicability of Machine Learning (ML) systems in real-world scenarios. Deep Learning (DL) models are now extensively used in critical domains such as medicine, transportation, remote sensing, and robotics, where the consequences of erroneous decisions can be severe. Consequently, it is vital for DNNs to provide reliable confidence scores that accurately quantify the true likelihood of their predictions, thus properly estimating their predictive uncertainty. For this reason, the problem of uncertainty calibration  (also referred as \textit{confidence calibration} \cite{kuppers2020multivariate} or simply \textit{calibration} \cite{guo2017calibration}) is becoming ubiquitous when developing DL models that are reliable and robust enough for real-world applicability.\\
\indent The importance of uncertainty calibration of DNNs is illustrated by the growing body of scientific work developed in recent years regarding this subject \cite{guo2017calibration,moon2020confidence,ovadia2019can,tian2021geometric,kull2019beyond,widmann2019calibration}. Nonetheless, most of these works are developed around the problem of uncertainty calibration in classification problems. Contrastingly, a significant number of DL safety-critical applications are related to object detection problems (\eg, autonomous driving, human-robot interaction, surveillance). We argue that one of the main reasons for a lack of scientific work regarding uncertainty calibration in object detection scenarios is due to the fact that there is no complete/proper theoretical and practical formulation regarding the understanding and evaluation of this problem. As such, this work aims at filling this gap, by proposing three uncertainty calibration evaluation metrics for object detection, based on a comprehensive theoretical framework (introduced in Section \ref{Definitions}). This framework focuses on \textit{semantic/label} uncertainty (instead of \textit{spatial} uncertainty, like other approaches to evaluate the probabilistic quality of detections \cite{hall2020probabilistic,kuppers2022parametric}), by leveraging \textit{Intersection Over Union} (IoU) in a threshold-based evaluation, akin to the conventional \textit{Mean Average Precision} (mAP). For this reason, our formulation for the problem of uncertainty calibration is consistent with the classical conception of an object detection problem. Additionally, and like in a standard evaluation framework for object detection, the metrics proposed in Section \ref{evaluation:metrics} can also have a stronger focus on localization performance, by averaging over different IoUs (\textit{e.g.}, from 0.5 to 0.95 in intervals of 0.05).  \\
\indent The \textbf{key contributions} of this work are twofold:
    \textbf{1)} A comprehensive theoretical formulation of the uncertainty calibration problem in object detection (Section \ref{Definitions}) that, to the best of our knowledge, is absent in relevant literature.
    \textbf{2)} Three novel uncertainty calibration metrics \footnote{Code for the proposed uncertainty calibration metrics available in the Supplementary Material.} specifically designed for the context of object detection evaluation (Section \ref{evaluation:metrics}), consistent with the mentioned theoretical formulation.

\section{Related Work}

The problem of uncertainty calibration is introduced to the DL community through the work presented in \cite{guo2017calibration}, where the calibration quality of various modern DNNs is evaluated on classification problems, with distinct datasets, from computer vision and natural language processing domains. The authors argue that despite their increased accuracy, modern DNNs suffer from significant miscalibration issues, which surpass those observed in `older' but less accurate architectures.\\
\indent The evaluation of uncertainty calibration in classification problems often relies on the widely used Expected Calibration Error (ECE) \cite{naeini2015obtaining}. However, limitations of this metric have been acknowledged in relevant literature \cite{nixon2019measuring, widmann2019calibration}. On the other hand, the use of \textit{proper scoring rules} \cite{gneiting2007strictly} like the Brier score \cite{brier1950verification} has been an increasingly common practice in recent literature regarding uncertainty calibration for classification problems \cite{ovadia2019can, tian2021geometric, moon2020confidence}.\\
\indent The lack of evaluation metrics specifically designed to the problem of uncertainty calibration in object detection is addressed in \cite{kuppers2020multivariate}, that proposes the Detection Expected Calibration Error (D-ECE). Since then, this metric has been adopted as a ``go-to" evaluation metric in different works regarding uncertainty calibration in object detection \cite{munir2022towards, pathiraja2023multiclass, munir2023bridging}. Comparably to the metrics proposed in this work, the D-ECE leverages the IoU for a threshold-based evaluation, thus being also focused on semantic uncertainty. Nonetheless, D-ECE is built on an incomplete formulation of the problem of uncertainty calibration in object detection, and therefore does not incorporate the effect of False Negative detections (see Section \ref{evaluation:metrics} for further details), which can be critical in safety-related applications.\\
\indent Other approaches to assess the reliability and probabilistic quality of object detector's predictions were proposed in recent years. The authors in \cite{hall2020probabilistic} propose Probability-based Detection Quality measure (PDQ), focusing on both \textit{label} and \textit{spatial} probabilistic quality; because of the focus on \textit{spatial} quality, ``PDQ has been primarily developed to evaluate \textit{new} types of probabilistic object detectors that are designed to quantify spatial and semantic uncertainties", which is not the case for most common state-of-the art object detectors like YOLO \cite{redmon2016you}, Fast R-CNN \cite{girshick2015fast} or SSD \cite{liu2016ssd}. The work in \cite{kuppers2022parametric} focuses specifically on spatial uncertainty and therefore proposes evaluating uncertainty calibration for object detection as a probabilistic regression task, by evaluating only object detectors with probabilistic regression output ``where a mean and a variance score is inferred for each bounding box quantity" \cite{he2019bounding}. Also, the authors in \cite{oksuz2023towards} propose the Self Aware Object Detection (SAOD) task, a testing framework which includes uncertainty calibration evaluation; for this evaluation, Localization Aware ECE (LaECE) is introduced, through the inclusion of spatial evaluation into the existing D-ECE \cite{kuppers2020multivariate}, by multiplying \textit{precision} with IoU; nonetheless, and similarly to the D-ECE, this metric is not affected by the existence of False Negatives.

\section{Uncertainty Calibration for Object Detection}
\label{Definitions}

In this section we formally introduce the concept of uncertainty calibration for the object detection domain. Since the proposed theoretical framework focuses on semantic uncertainty, the presented definitions are inspired in the concept of uncertainty calibration for classification problems \cite{widmann2019calibration}, though adapted to the particularities of the object detection's case.   \\
\indent For the subsequent definitions we will consider $\mathcal{X}$, the sample space of inputs, $(\mathcal{Y},\mathcal{K})$ the sample space of bounding-box locations and associated classes. We can now consider the respective random variables, $X$ and $(Y,K)$, defined under those sample spaces. Let us note that a realization of $(Y,K)$ can define an arbitrary number of locations (bounded by the total number of possible locations) and respective classes \ie, any realization of $(Y,K)$, for a problem with $C$ different classes, is a subset of $\Omega \times \{1,\ldots,C\}$, where $\Omega \subset \mathbb{N}^4$. For this reason, we will denote as $P\big((y,c) \in (Y,K) \big)$ the abbreviation of $P\big(\{ \vartheta, \psi \in (\mathcal{Y},\mathcal{K}):  y \in Y(\vartheta) \wedge c \in K(\psi) \} \big)$ \ie, the probability that some singular bounding-box location (with respective class) belongs to some realization of $(Y,K)$.\\
\indent Before considering the concept of calibration in the context of object detection, we have to define the mathematical object for which such considerations are pertinent, \ie, a \textbf{confidence-based detector}. Let us first consider the function $\Psi: \mathcal{X} \rightarrow (\mathcal{Y},\mathcal{K})$, that maps an input from the sample space $\mathcal{X}$ to the set of all possible bounding-box locations in that specific input. We are now in condition to assert the following definition.

\begin{defi}
\label{probabilistic:detector}
    Let us define a \textbf{confidence-based detector} (for a problem with $C$ classes) as a function $f: \mathcal{X} \rightarrow \Phi$ that maps each input to a set of all possible bounding-box locations and correspondent confidence scores, for each respective class.  Formally, $\forall x \in \mathcal{X}$ we have that
    \begin{align}
         & \textit{1}. \quad \ f(x) \subseteq \Omega \times \{1,\ldots,C\} \times [0,1]  \\ 
         & \textit{2}. \quad \ (l,c) \in \Psi(x) \iff (l,c,p) \in f(x) .
    \end{align}
\end{defi}
\medskip
We note that $(l,c,p) \in f(x)$ denotes a bounding-box detection of the form (\textit{Location, Class, Confidence score}).\\
\indent Although in a practical scenario most object detection systems will not consider all possible bounding-box locations, from a theoretical point-of-view Definition \ref{probabilistic:detector} is a reasonable definition, because we can associate all disregarded locations with a confidence score equal to 0. As such, this definition is consistent with common state-of-the-art object detectors. \\
\indent Let us observe that, since most object detection systems incorporate suppression strategies (like \textit{Non Maximum Suppression}), considering the concept of calibration for precise localization can be unreasonable in a practical scenario. Therefore, let us take as 
$ \mathcal{F}: \Omega \times \Omega \rightarrow [0,1] $ to designate the well known IoU function, and define (for some threshold value $\tau \in [0,1]$) $\varphi_\tau: \Omega \times \{1,\ldots,C\} \rightarrow [0,1] $ as
\begin{align}
    \varphi_{\tau}(y,c) = \max \big{\{} p  : \ (l,c,p) \in f(X) 
      %&  \wedge c=k \notag \\
        \wedge \mathcal{F}(y,l) \geq\tau \big{\}},
\end{align}
%, for $(l,c,p) \in f(x)$ and $y$ a ground-truth bounding-box location.
referring to the maximum confidence value, for a given class, under the IoU threshold conditions. It is worth observing that, following the definition of a \textit{confidence-based detector} (Definition \ref{probabilistic:detector}), the value of $\varphi(y,c)$ can be 0 (this is the case were there is no positive prediction that satisfies the condition  $\mathcal{F}(y,l) \geq\tau$). We can now consider the following definition.

% % % % DEFINITION 2
\begin{defi}
\label{global:calibration:detection}
    We say that a confidence-based detector $f: \mathcal{X} \rightarrow \Phi$ is \textbf{globally calibrated} (for some threshold value $\tau \in [0,1]$) iff 
    \begin{align}
    \label{eq:global:calib}
        P\big( (y,c) \in (Y,K) \ | \ (l,c,p) \in f(X)  \big) = \varphi_{\tau}(y,c).   
    \end{align}
   
\end{defi}
Based on Definition \ref{global:calibration:detection} we will develop new evaluation metrics with the purpose of assessing the calibration of the predictive uncertainty of confidence-based detectors in a practical object detection scenario.\\
\indent Definition \ref{global:calibration:detection} requires that, for every portion of the input, there is a spatial bounding box neighborhood (defined under IoU conditions) whose confidence value (that can be 0) correctly codifies the likelihood of the existence of a certain object. On the other hand, a weaker (and therefore incomplete) formulation of the uncertainty calibration problem in object detection (similar to that used in \cite{kuppers2020multivariate}) can also be defined, by considering only the probabilistic quality of the detections that have a positive confidence score (\ie, the detections that are actually returned by a model in a practical scenario). The latter is outlined below. 

\begin{defi}
\label{local:calibration:detection}
    We say that a confidence-based detector $f: \mathcal{X} \rightarrow \Phi$ is \textbf{locally calibrated} (for some threshold value $\tau \in [0,1]$) iff
    \begin{align}
    \label{eq:local:calib}
        P\big((y,c) \in (Y,K) \ | \   (l,c,p) \in f(X),  %\notag 
        \ p>0 , \  \mathcal{F}(y,l) \geq\tau  \big)  = \varphi_{\tau}(y,c).   
    \end{align}
\end{defi}
We care to note that such formulation does not take into account the existence of False Negatives.

\section{Evaluation Metrics}
\label{evaluation:metrics}
%\paragraph{Notation:} We define \textit{bag} (also called a \textit{multiset}) as an extension of the notion of \textit{set}, that can have repeated elements (\ie, different instances of the same element).\\
\textit{Note}: For the remainder of this work we define \textit{bag} (also called a \textit{multiset}) as an extension of the notion of \textit{set}, that can have repeated elements (\ie, different instances of the same element).\\
\indent Evaluating uncertainty calibration in a practical object detection setting encompasses some of the same challenges found when evaluating this problem in a classification scenario, regarding the nonexistence of ground-truth information for the true likelihood values (left side of both Equations \eqref{eq:global:calib} and \eqref{eq:local:calib}). For this reason, arises the need to develop practical evaluation metrics in relation to the previously outlined formal definitions. Therefore, considering the theoretical formulation presented in Definition \ref{global:calibration:detection}, we have developed uncertainty calibration evaluation metrics in the context of object detection.\\
\indent Before introducing such metrics, some concepts have to be outlined, that are common in the context of object detection problems and will be necessary when evaluating uncertainty calibration. Let us take a confidence-based detector $f: \mathcal{X} \rightarrow \Phi$, a finite set of inputs $\mathcal{X}' \subset \mathcal{X}$, and the bag of respective ground truth locations and classes $(\mathcal{Y}',\mathcal{K}') \subset (\mathcal{Y},\mathcal{K})$. In a setting of this type, and for some threshold value $\tau \in [0,1]$, we can define: $\textit{TP}_{\tau}$, as the bag of confidence scores associated with \textit{True Positives} \ie, detections that have a corresponding ground-truth (with an IoU equal or greater than $\tau$); $\textit{FP}_\tau$ as the bag of confidence scores associated with \textit{False Positives} \ie, detections that do not have a corresponding ground-truth; and $\textit{FN}_\tau$ as the bag of confidence scores associated with \textit{False Negatives} \ie, ground-truth bounding boxes with no corresponding detection. Although $\textit{FN}_\tau$ is a bag of identical (theoretically zero-valued) confidence scores, the number of such detections will be fundamental in the computation of the proposed uncertainty calibration metrics.\\
\indent In the following subsections we propose three novel uncertainty calibration metrics based on Definition \ref{global:calibration:detection}. The first two metrics (Subsections \ref{qgc} and \ref{sgc}) are based on \textit{proper scoring rules} \cite{gneiting2007strictly}, while the third one (Subsection \ref{Egce}) relies on bin-wise computations, similarly to the ECE \cite{naeini2015obtaining}.

\subsection{Quadratic Global Calibration Score}
\label{qgc}

Inspired in the Brier score \cite{brier1950verification} - a \textit{proper scoring rule} widely used to evaluate uncertainty calibration in classification problems - we introduce in this section the Quadratic Global Calibration score (QGC), that leverages the same fundamental principle of its predecessor by computing the quadratic difference between a confidence score and its true response. \\
\indent We start by considering a confidence-based detector  $f: \mathcal{X} \rightarrow \Phi$, a finite set of inputs $\mathcal{X}'$ and the bag of respective ground truth locations and classes $(\mathcal{Y}',\mathcal{K}')$. In this context we can construct our bags $\textit{TP}_\tau$, $\textit{FP}_\tau$ and $\textit{FN}_\tau$. The QGC can now be computed as
\begin{align}
\label{qlcs:conventional:1}
    \text{QGC} & = \sum_{p \in \textit{TP}_\tau \cup \textit{FN}_\tau} (p-1)^2 +  \sum_{p \in \textit{FP}_\tau} p^2 \\
     &  = \sum_{p \in \textit{TP}_\tau} (p-1)^2 +  \sum_{p \in \textit{FP}_\tau} p^2 + |\textit{FN}_\tau|.
\end{align}
We remind that $p$ denotes a confidence score. A lower value of QGC translates a better performance in terms of uncertainty calibration, reaching its optimal value at 0.

\subsection{Spherical Global Calibration Score}
\label{sgc}

The Spherical Global Calibration score (SGC) is inspired in a less common \textit{proper scoring rule}, the Spherical score \cite{gneiting2007strictly}. Let us first denote
\begin{align}
    r(p)=\sqrt{p^2+(1-p)^2}.
\end{align}
A direct adaptation of the Spherical score would be formulated as 
\begin{align}
    SGC^* = \sum_{p \in \textit{TP}_\tau \cup \textit{FN}_\tau} \frac{p}{r(p)} + \sum_{p \in \textit{FP}_\tau} \frac{1-p}{r(p)}.
\end{align}
Such formulation reaches its optimal value at $N=|\textit{TP}_\tau|+|\textit{FP}_\tau|+|\textit{FN}_\tau|$, while a higher value translates a better performance (similar to the original Spherical score). As such, we define the SGC as
\begin{align}
    SGC 
    %& = \sum_{p \in \textit{TP}_\tau \cup \textit{FN}_\tau} \frac{r(p)-p}{r(p)} + \sum_{p \in \textit{FP}_\tau} \frac{r(p)-p+1}{r(p)} \\
    & =  \sum_{p \in \textit{TP}_\tau \cup \textit{FN}_\tau} \left(1 - \frac{p}{r(p)}\right) + \sum_{p \in \textit{FP}_\tau} \left(1 - \frac{1-p}{r(p)}\right) \\
    & = N - \sum_{p \in \textit{TP}_\tau} \frac{p}{r(p)} - \sum_{p \in \textit{FP}_\tau} \frac{1-p}{r(p)}.
\end{align}
A lower value of SGC translates a better performance in terms of uncertainty calibration, reaching its optimal value at 0.

\PC{
\begin{table*}[t]
    \centering
    \begin{adjustbox}{max width=0.9\textwidth}
    \begin{tabular}{c| c c c c c }
    \toprule

    ~ & mAP $\uparrow$ & QGC $\downarrow$ & SGC $\downarrow$ & EGCE $\downarrow$ & D-ECE $\downarrow$\\

    \hline\hline \rule{0pt}{3ex}

    YOLOv5 (Nano) & 0.44 & 23952.82 & 24882.37 & 21373.61 & 3626.55 \\
    \rule{0pt}{3ex}
    YOLOv5 (Small) & 0.53 & 20950.61 & 21857.65 & 17188.79 & 2883.91 \\
    \rule{0pt}{3ex}
    YOLOv5 (Medium) & 0.65 & 18477.24 & 19390.43 & 13461.12 & 2194.51 \\
    \rule{0pt}{3ex}
    YOLOv5 (Large) & 0.67 & 17578.47 & 18490.01 & 11831.19 & 1624.24 \\
    \rule{0pt}{3ex}
    YOLOv5 (Extra Large) & \textbf{0.69} & \textbf{17085.51} & \textbf{17987.22} & \textbf{11100.42} & \textbf{1305.86} \\
    
    \bottomrule
    \end{tabular}
    \end{adjustbox}
    \caption{Comparing the performance of five different variations of the YOLOv5 \cite{glenn_jocher_2022_7347926} object detector, evaluated with mAP, QGC, SGC, EGCE and D-ECE, using an IoU threshold of 0.5.}
    \label{comparison:1}
\end{table*}
}

\subsection{Expected Global Calibration Error}
\label{Egce}

The Expected Global Calibration Error (EGCE) is an adaptation of the popular ECE \cite{naeini2015obtaining}. The ECE is widely used to evaluate uncertainty calibration in classification problems, and works based on the principle of computing the bin-wise difference between average confidence scores and average accuracy. Since the original ECE leverages the concept of ``accuracy" - common in the evaluation of classification systems - the use of the EGCE will require an adaptation of this concept to the context of object detection. In fact, such challenge is also addressed in \cite{kuppers2020multivariate} and is reflected on the development of the D-ECE. Therefore, EGCE will be based on the same principle of the D-ECE, but incorporating the necessary adaptations to address the previously discussed limitations of its counterpart.\\
\indent We start by creating the sets of bins $\{B_1^{\textit{TP}_\tau}, \ldots, B_M^{\textit{TP}_\tau} \}$ and $\{B_1^{\textit{FP}_\tau}, \ldots, B_M^{\textit{FP}_\tau} \}$, where each bin is a bag of confidence scores defined as 
\begin{align}
    %& B_1 =  \big{\{} p_j : (l_j,p_j) \in \mathcal{I} \ \wedge p_j \in [0,1/M] \big{\}}, \\
    B_i^{\textit{TP}_\tau} = \big{\{} p \in \textit{TP}_\tau :  p \in \ ](i-1)/M,i/M] \big{\}},\\
     B_i^{\textit{FP}_\tau} = \big{\{} p \in \textit{FP}_\tau :  p \in \ ](i-1)/M,i/M] \big{\}},
\end{align}
for $i=1,\ldots,M$. Additionally, we can now consider the set of bins $\{B_1,\ldots,B_M\}$ where each bin is a bag defined as
$
    B_i = B_i^{\textit{TP}_\tau} \cup B_i^{\textit{FP}_\tau},
$
for $i=1,\ldots,M$. For each bin, we define the \textit{average confidence} and the \textit{precision} per bin, respectively, as
\begin{align}
    \text{conf}(B_i) = \frac{1}{|B_i|} \sum_{p \in B_i} p, \qquad
    \text{prec}(B_i) = \frac{|B_i^{\textit{TP}_\tau}|}{|B_i^{\textit{TP}_\tau}| + |B_i^{\textit{FP}_\tau}|}.
\end{align}
We can now outline the definition of the D-ECE as 
\begin{align}
\label{d-ece}
    \text{D-ECE} = \sum_{i=1}^M |B_i| \left|\text{prec}(B_i) - \text{conf}(B_i)\right|.
\end{align}
\textit{Note:} in fact, the definition given in \cite{kuppers2020multivariate} is the average of \eqref{d-ece} \ie, divided by $|\textit{TP}_\tau \cup \textit{FP}_\tau|$.\\
\indent Because D-ECE evaluates only the calibration of the detections that have a positive confidence score (\ie, the bags $\textit{TP}_\tau$ and $\textit{FP}_\tau$), it can be considered a metric for \textit{local} calibration (Definition \ref{local:calibration:detection}).\\
\indent The EGCE will leverage the D-ECE's principle of contrasting \textit{precision} and \textit{confidence}. Nonetheless, contrarily to the latter, the EGCE will incorporate False Negative detections by considering them analogous to False Positives with a confidence of 1, and therefore being incorporated in the last bin. As such, let us start by considering
\begin{align}
    \delta (B_i) = \frac{|B_i^{\textit{TP}_\tau}|}{|B_i^{\textit{TP}_\tau}| + |B_i^{\textit{FP}_\tau}| + |\textit{FN}_\tau|}.
\end{align}
We can now finally define the EGCE as
\begin{align}
\label{egce}
    \text{EGCE} = \sum_{i=1}^{M-1} |B_i| \left|\text{prec}(B_i) - \text{conf}(B_i)\right| 
     + |B_M| \left|  \delta (B_M) - \text{conf}(B_M) \right|.
\end{align}
Following similar guidelines of those in \cite{kuppers2020multivariate}, we will only consider detections with a confidence value above a given threshold (in our case 0.1), when constructing the bags $\textit{TP}_\tau$, $\textit{FP}_\tau$ and $\textit{FN}_\tau$ for calculating the D-ECE and EGCE (this avoids a bias to the behaviour of low-confidence detections, that is common in this type of bin-wise metrics).  Both the D-ECE and the EGCE represent a better performance (in terms of uncertainty calibrations) with lower values, reaching their optimal value at 0. The common number of bins for these types of metrics is between 10 and 20 bins, thus, in our experiments, we will use 15 bins.

% % % % New Section ======================
\section{Experiments and Results}

\textit{Note}: For readability purposes, we will subsequently use the acronyms for True Positive (TP), False Positive (FP) and False Negative (FN), with no confusion with the associated mathematical objects $\textit{TP}_\tau$, $\textit{FP}_\tau$ and $\textit{FN}_\tau$.\\
\indent The experiments described hereafter have been done with YOLOv5 \cite{glenn_jocher_2022_7347926} object detectors. When using the COCO dataset \cite{cocodataset}, the pre-trained models provided by the YOLOv5 developers \cite{glenn_jocher_2022_7347926} have been evaluated on the COCO validation set with 5000 images (since the official test set has no available ground-truth). When using the PASCAL VOC (2012) \cite{pascal-voc-2012} dataset, the models are trained for 100 epochs starting with random weights and standard YOLOv5 hyper-parameters; the available PASCAL VOC data is divided randomly into training and test sets, with a 70/30 split. A variety of representative experiments, designed to give a wide understanding on how the proposed metrics behave under various circumstances, have been performed and the results are compared to the existing D-ECE metric. Each subsection encapsulates one ore more pertinent scientific questions, as summarized below. \\
\indent \textbf{\textit{Calibration} vs. \textit{performance}} (Subsection \ref{results_general}): How do the uncertainty calibration metrics behave when evaluating deep models with increasing mAP performance under distinct IoU threshold conditions. \textbf{Sensitivity tests} (Subsection \ref{results_main}.): How do the metrics react to increasing proportions of specific types of detections (\eg, FNs, high-confidence FPs, low-confidence TPs) and what specific properties can be derived from their behaviour. \textbf{Effects of distribution-shifts and calibration strategies} (Subsection \ref{results_shift}): How does the introduction of distribution-shifts on the test set impact the evaluation done with the employed metrics; how do state-of-the-art strategies - that improve uncertainty calibration in classification problems - work in the context of these metrics, when adapted to object detection.\\

\subsection{\textit{Calibration} vs. \textit{performance}}
\label{results_general}

\begin{figure}[t]

    \includegraphics[width=1\linewidth]{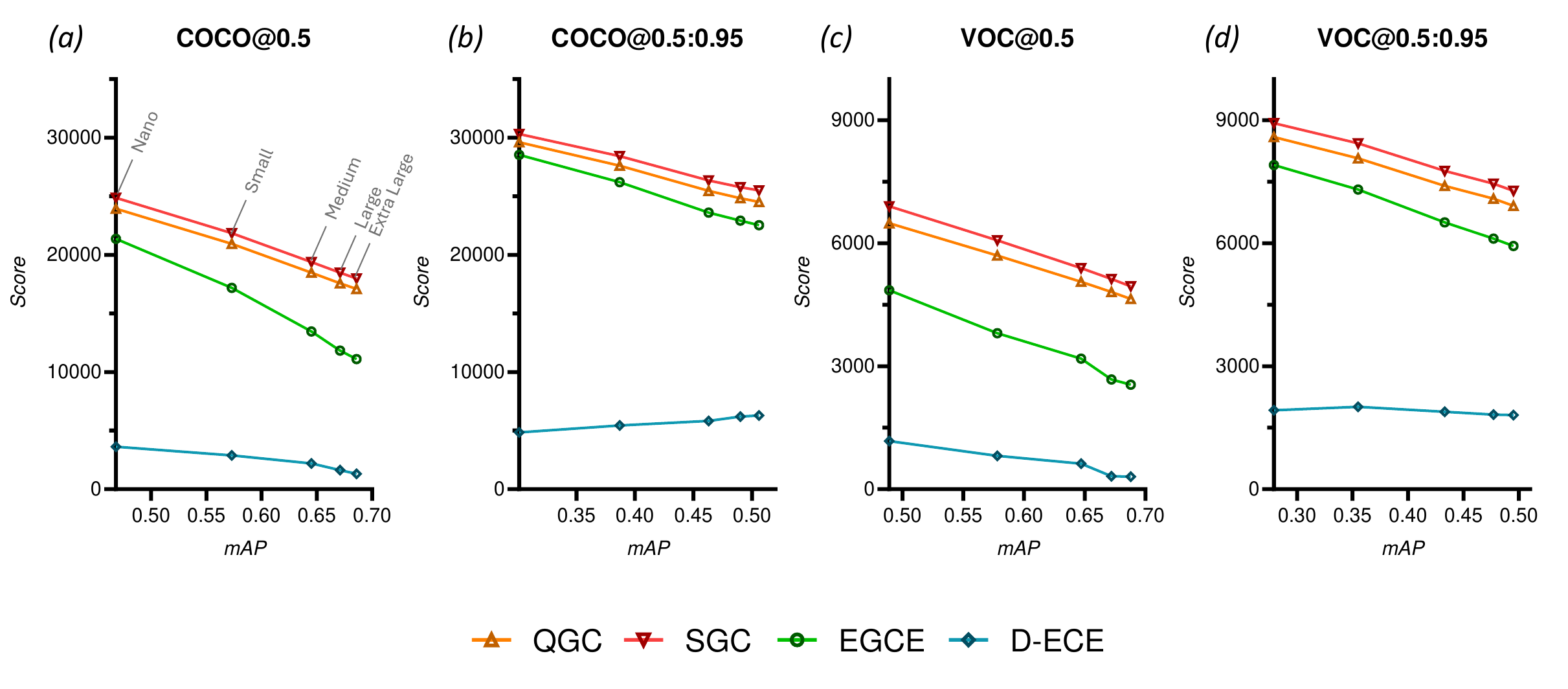}  
    \caption{Evaluating QGC, SGC, EGCE and D-ECE - against mAP - using YOLOv5 models (\textit{Nano}, \textit{Small}, \textit{Medium}, \textit{Large} and \textit{Extra Large}). mAP increases proportionally to the model capacity shown from left to right (highlighted as grey-text legend in \textbf{(a)}). The results were obtained: on the COCO dataset with \textbf{a)} IoU threshold of 0.5, \textbf{b)} averaging the results with IoU threshold values between 0.5 and 0.95, with a step of 0.05; on the PASCAL VOC dataset with \textbf{c)} IoU threshold of 0.5; \textbf{d)} averaging the results with IoU threshold values between 0.5 and 0.95, with a step of 0.05.}
    \label{fig_mAP}
\end{figure}

\indent Figure \ref{fig_mAP} shows the performance of the different versions of the YOLOv5 object detector, namely \textit{Nano}, \textit{Small}, \textit{Medium}, \textit{Large} and \textit{Extra Large} (respectively with 3.2, 12.6, 35.7, 76.8 and 140.7 million parameters), when evaluated using the  proposed metrics (QGC, SGC, EGCE) and the D-ECE (for comparison) against the classical mAP evaluation. The values of the uncertainty calibration metrics are presented with absolute values (instead of the average) because, unlike what happens in classification problems (where is common to present the averaged values), distinct object detection models can output a varying number of detections. As such, averaging could create problems when comparing the performance of different models; \eg, a model that outputs a large number of low-confidence FP detections could appear to perform better in terms of uncertainty calibration than a model with a relatively lower number of FP detections, because of proportionality issues. Additionally, we care to note that it is expectable to achieve an absolute value of the D-ECE smaller than the QGC, SGC, and EGCE, because the D-ECE metric does not incorporate FN detections.\\
\indent From Fig. \ref{fig_mAP} we can infer some key observations. When considering the evaluation with an IoU threshold of 0.5 (Figures \ref{fig_mAP}.a, \ref{fig_mAP}.c), there is a relationship between the uncertainty calibration metrics and mAP, with the models showing better calibration results (\ie, lower scores) as the mAP performance improves; this relation is stronger with the proposed metrics (QGC, SGC, EGCE) than with the D-ECE. Considering the cases where we average the results from different IoU thresholds (Figures \ref{fig_mAP}.b, \ref{fig_mAP}.d), similar conclusions can be made for the proposed metrics but not for the D-ECE, where the latter is progressively aggravated with better performing models (Figure \ref{fig_mAP}.b), or shows an inconsistent relation to mAP evaluation (Figure \ref{fig_mAP}.d). A deeper look into the reason behind this phenomenon is taken in Supplementary Material by analysing the evolution of the scores with increasing IoU threshold values, for the \textit{Nano} and \textit{Extra Large} versions of Yolov5.\\
\indent The \textbf{main take}, from the results reported in this Subsection, is that the three proposed uncertainty calibration metrics show a relation to mAP performance evaluation that is robust to different IoU conditions. Nonetheless, it is observable that strong improvements in performance do not translate proportionally to uncertainty calibration evaluation.

\subsection{Sensitivity tests}
\label{results_main}

\begin{figure}[h!]

    \includegraphics[width=1\linewidth]{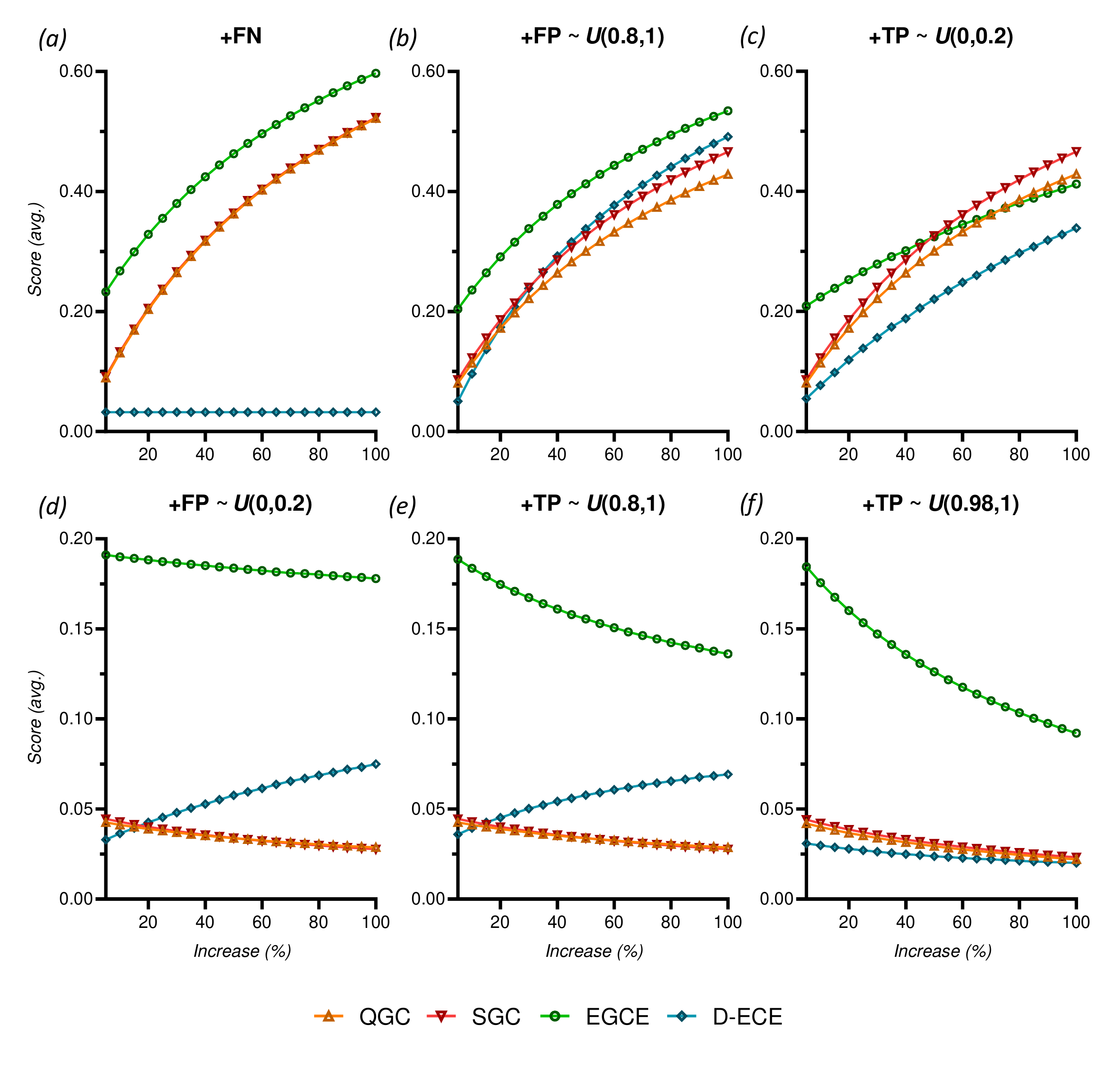}  
    \caption{Evaluating QGC, SGC, EGCE and D-ECE for increasing proportions of: \textbf{a)} FN detections; \textbf{b)} FP detections with confidence scores extracted from the Uniform distribution $U[0.8,1]$; \textbf{c)} TP detections with confidence scores extracted from $U[0,0.2]$; \textbf{d)} FP  detections with confidence scores extracted from $U[0,0.2]$; \textbf{e)} TP detections with confidence scores extracted from $U[0.8,1]$; \textbf{f)} TP detections with confidence scores extracted from $U[0.98,1]$.}
    \label{fig_main}
\end{figure}

\indent Figure \ref{fig_main} portraits a series of results that evaluate how different types of detections influence the uncertainty calibration metrics (QGC, SGC, EGCE and D-ECE). As an example, in Figure \ref{fig_main}.b we gradually increase the number of FPs with confidence scores extracted from a Uniform distribution in the interval [0.8,1]; for instance, an \textit{increase} of 60\% means that it has been added a number of such detections equal to 60\% of the original number of detections (\ie, $0.6 \times (|\textit{TP}_\tau|+|\textit{FP}_\tau|+|\textit{FN}_\tau|)$). Specifically, we have carried out these experiments with FNs (Figure \ref{fig_main}.a), low-confidence and high-confidence FPs (Figures \ref{fig_main}.b and \ref{fig_main}.d) and also low-confidence and high-confidence TPs (Figures \ref{fig_main}.c, \ref{fig_main}.e and \ref{fig_main}.f); further details can be found in the caption of Figure \ref{fig_main}. In these experiments the ``starting point" results are based on the COCO pre-trained YOLOv5 (\textit{Large}). The results are presented from 5\% until 100\% increase, with a step of 5\%. The values of the previously referred metrics are averaged (contrarily to when we were comparing different models) because in this situation we are actually interested in analysing how specific types of detections proportionally influence the uncertainty calibration metrics. The IoU threshold is set at 0.5.\\
\indent It is important to observe that, in the context of uncertainty calibration, it is expectable that evaluation metrics penalise both overconfident and underconfident detections, specifically: overconfident FPs (Figure \ref{fig_main}.b), underconfident TPs (Figure \ref{fig_main}.c) and also FNs (Figure \ref{fig_main}.a). On the other hand,  the metrics are expected to reward higher proportions of high-confidence TPs (Figs \ref{fig_main}.e and \ref{fig_main}.f), and even low-confidence FPs (Fig. \ref{fig_main}.d).\\
\indent We start by comparing the behaviour of the \textit{proper scoring rule}-based metrics, QGC and SGC. It can be observed that the metrics present a similar behaviour in most cases; specifically, in Figures \ref{fig_main}.e, \ref{fig_main}.d, \ref{fig_main}.f and \ref{fig_main}.a their behaviour is illustrated as nearly identical. In Figures \ref{fig_main}.b and \ref{fig_main}.c, although still similar, we observe that the SGC presents a more sensitive behaviour (reflected by stronger increase in value) than the QGC.\\
\indent We can now compare the behaviour of the EGCE with the QGC and SGC. We start by observing that QGC and SGC behave in a symmetrical way, meaning that adding high-confidence FPs  produces the same negative effect as adding low-confidence TPs (comparing Figs \ref{fig_main}.b and \ref{fig_main}.c), just like adding high-confidence TPs produces the same positive effect as adding low-confidence FPs (comparing Figures \ref{fig_main}.e and \ref{fig_main}.d). Contrarily, the EGCE penalizes high-confidence FNs more than low-confidence TPs, as well as rewards high-confidence TPs more favorably than low-confidence FPs. This behaviour can be advantageous since it is more consistent with a general evaluation of an object detection model (that will obviously favour TP detections over the FP counterparts).\\
\indent Finally, we compare the behaviour of the three uncertainty calibration metrics proposed in this work (QGC, SGC, EGCE) against D-ECE. As previously referred, the D-ECE can be interpreted through our theoretical formulation as a \textit{local} calibration metric (in contrast to the proposed \textit{global} calibration metrics); therefore, as illustrated in Figure \ref{fig_main}.a, the D-ECE has no sensibility to FNs, while the other metrics show a significant decrease in performance when exposed to increasing proportions of FNs. Furthermore, contrarily to the other metrics, the D-ECE still has small increases in values when exposed to a larger proportion of supposedly ``desirable" detections (\ie, high-confidence TPs and low-confidence FPs); however, in the edge case presented in Fig. \ref{fig_main}.f we witness a slight decrease in D-ECE. \\
\indent Finally, the \textbf{main conclusions} are: contrarily to the D-ECE, the metrics QGC, SGC, and EGCE show robust sensitivity in the presence of FNs and behave as expected with increasing proportions of ``desirable" detections; the QGC and the SGC show similar behaviour, with the SGC being sightly more sensitive to increasing proportions of ``undesirable" detections; the EGCE, contrarily to both the QGC and SGC, does not have a symmetrical behaviour towards TPs and FPs, and also shows higher sensibility with increasing proportions of ``desirable" detections than the latter referred metrics.

\subsection{Effects of distribution-shifts and calibration strategies}
\label{results_shift}

\begin{figure}[t]

    \includegraphics[width=1\linewidth]{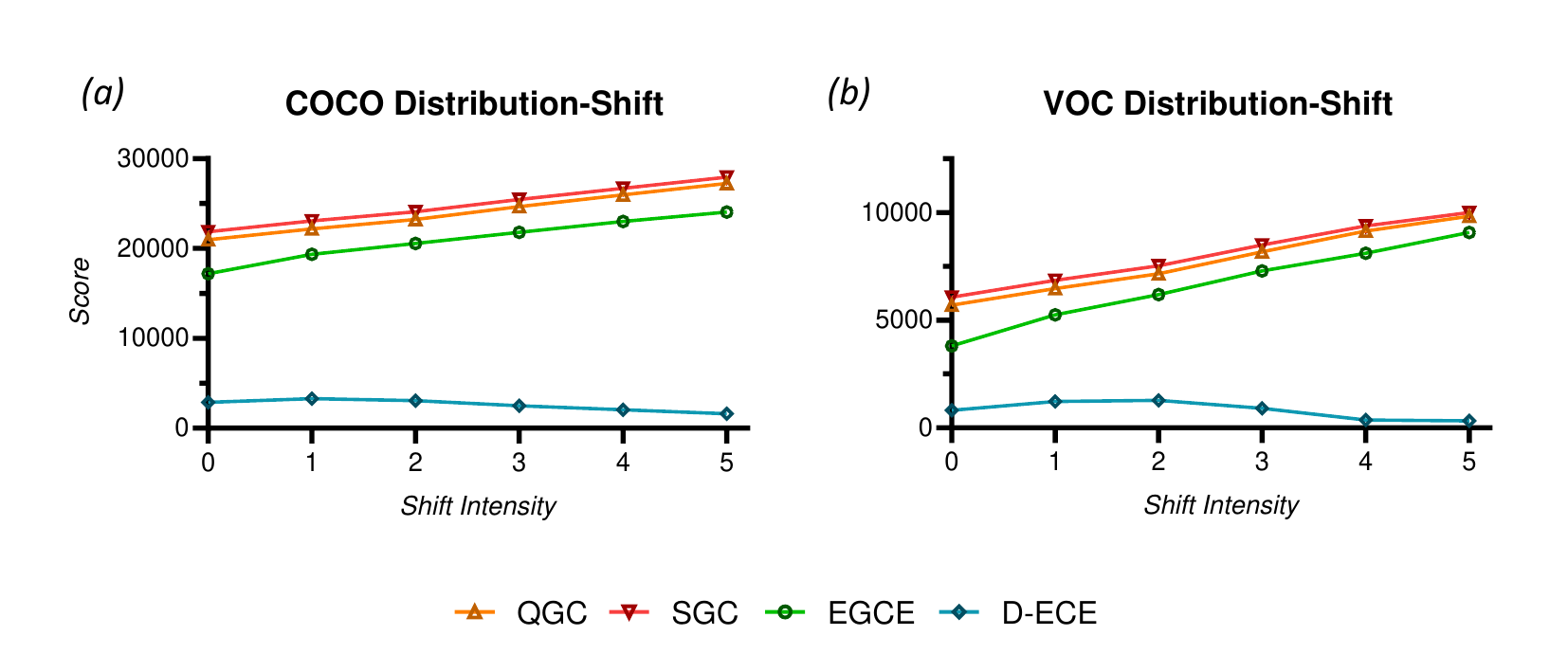}  
    \caption{Evaluating QGC, SGC, EGCE and D-ECE, with increasing intensity of shifts in the distribution of the test data, using Yolov5 (\textit{Small}) with \textbf{a)} the COCO dataset and \textbf{b)} the PASCAL VOC dataset.}
    \label{fig_shift}
\end{figure}

\begin{figure}[h!]

    \includegraphics[width=1\linewidth]{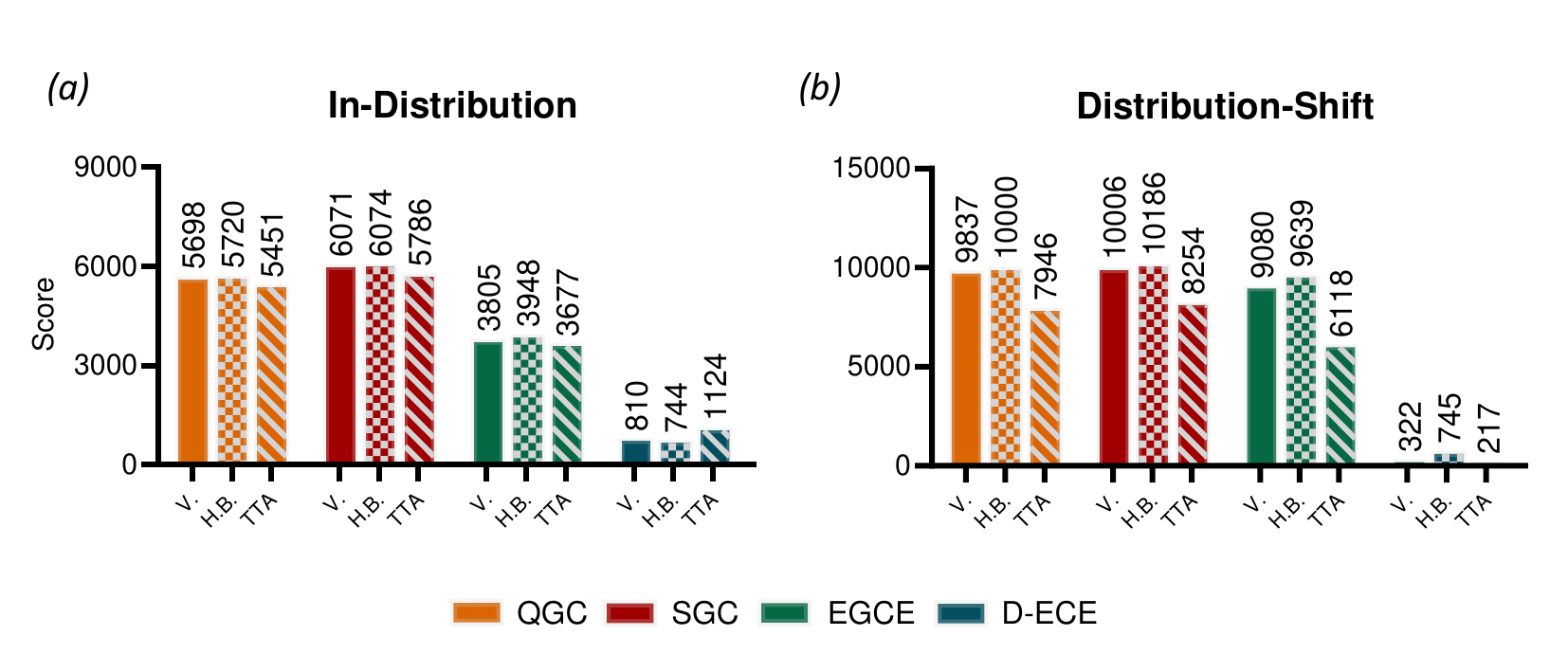}  
    \caption{Evaluating QGC, SGC, EGCE and D-ECE, after applying \textit{histogram binning} (H.B.) and TTA, compared to a \textit{vanilla} (V.) approach - \ie with no calibration strategy - using Yolov5 (\textit{Small}) in the PASCAL VOC test set \textbf{a)} with no distribution-shift and \textbf{b)} with level 5 distribution-shift.}
    \label{fig_methods}
\end{figure}

We start by evaluating the effect that distribution-shifts have in terms of uncertainty calibration, as quantified by the proposed metrics and the D-ECE (results shown in Fig. \ref{fig_shift}). These type of shifts induced on the test data have shown to induce negative effects in the uncertainty calibration of DNNs in classification scenarios \cite{ovadia2019can}. For the purpose of these experiments, and similarly to what is done in \cite{ovadia2019can}, the distribution-shifts are artificially induced with 5 different degrees of intensity (see details in Supplementary Material). \\
\indent Regarding the results, similar conclusions can be derived from both Figures \ref{fig_shift}.a and \ref{fig_shift}.b. First, all three proposed uncertainty calibration metrics demonstrate a consistent increase as the intensity of the shift rises; this is more evident on the PASCAL VOC dataset (Figure \ref{fig_shift}.b) where the scores almost double in value. This type of consistent increase is in line with what is observed with classical uncertainty calibration metrics used in classification problems \cite{ovadia2019can}. On the other hand, when looking at the D-ECE, the behaviour of this metric is inconsistent to what is expected, with small increases in low degrees of shift intensity, followed by a decrease in the score as the intensity of the shift is aggravated.\\
\indent The latter portion of this subsection is focused on a concise examination of how calibration strategies impact the uncertainty calibration metrics employed earlier. \textit{Histogram binning} \cite{zadrozny2001obtaining, kuppers2020multivariate} (code in Supplementary Material) and \textit{test time augmentation} (TTA) are the techniques chosen for these experiments (details on these techniques in Supplementary Material). The rationale behind this choice lies in a relevant fundamental difference between these two techniques: while \textit{histogram binning} only acts on the confidence value of existing detections (therefore only addressing \textit{local} calibration), TTA can also decrease the rate of FNs besides altering existing confidence values (possibly addressing \textit{global} calibration in the process). On an additional note, although not a common strategy, some evidence of the positive effects of TTA-based strategies in uncertainty calibration has already been outlined  \cite{Conde_2022_BMVC}.\\
\indent For the discussion of the results, we start by analysing Figure \ref{fig_methods}.a, where the experiments are made in the in-distribution (\ie, with no distribution-shift) PASCAL VOC test set. As expected, \textit{histogram binning} is not capable of improving the \textit{global} calibration metrics (QGC, SGC, EGCE) - improving only the D-ECE - while TTA induces small decreases in those metrics (but not in the D-ECE). Regarding the results in Figure \ref{fig_methods}.b, where the experiments are made with a shifted test set, we start by observing that \textit{histogram binning} is not capable of improving any uncertainty calibration metric; this is somewhat expected given that this technique relies on the distribution of its training data, which, in this case, differs from that of the test set. TTA shows relatively good improvements in terms of QGC, SGC and EGCE, and a small improvement in terms of D-ECE.\\
\indent The \textbf{main observations} are: contrarily to the D-ECE, the three proposed metrics show a behaviour under distribution-shifts that is congruent to what has been already observed in classification problems; there is evidence to suggest that typical \textit{post-hoc} calibration methods, that only alter the confidence value of existing detections, may not be sufficient in the context of \textit{global} calibration evaluation in object detection scenarios.

\section{Final Remarks}
This article introduces a comprehensive theoretical and practical framework for assessing uncertainty calibration in object detection. The conceptual distinction between \textit{global} and \textit{local} calibration, outlined in this work, proved to be not only useful as theoretical foundation for the development of new evaluation metrics, but also for understanding the underlying fundamental differences between these metrics and the existing D-ECE.\\
\indent The evaluation metrics proposed in the context of our framework are successfully used to evaluate the calibration of uncertainty estimates in various object detection scenarios. From these experiments, some interesting concluding remarks can be derived regarding some of the intrinsic properties of the proposed metrics that, in contrast, were not found in D-ECE: \textbf{1)} they show consistent relation to mAP evaluation under different IoU threshold conditions; \textbf{2)} they show robust sensitivity to varying proportions of representative types of detections; \textbf{3)} their response under distribution-shifts is in line with what has been observed in classification scenarios.\\
\indent Since it was outside the main scope of the article, this work has some limitations regarding the experimentation with different calibration strategies\PC{the experiments carried out in this work does not explore all calibration strategies but concentrates on the relevant ones}. Nonetheless, the presented evidence seems to suggest that, under a complete formulation for the problem of uncertainty calibration, there is a need to re-think the way calibration techniques are developed and applied, specially when considering \textit{post-hoc} strategies that only act on the confidence values of existing detections.\\
\indent On a final note, since the QGC and the SGC have a fairly similar behaviour as uncertainty calibration metrics, we suggest that the application of one of these metrics - paired with the EGCE - is sufficient for assessing the \textit{global} calibration of object detection systems.

\PC{
\begin{itemize}
    \item Contrarily to the D-ECE, the three proposed evaluation metrics (QGC, SGC, EGCE) show robust sensitivity in the presence of FNs. 
    \item The QGC and the SGC show similar behaviour, with the SGC being sightly more sensitive to increasing proportions of ``undesirable" detections. As such, choosing between only one of these metrics seems to be reasonable.
    \item The EGCE, contrarily to both the QGC and the SGC, does not have a symmetrical behaviour towards TPs and False Positives, and also shows higher sensibility with increasing proportions of ``desirable" detections, than the latter referred metrics. As such it is useful to use both EGCE and a proper scoring rule-based metric, since they offer different perspectives to the evaluation of uncertainty calibration.
    \item Although shown to be less robust than the proposed metrics, the D-ECE can still be useful to strictly assess \textit{local calibration} (instead of \textit{global calibration}). 
\end{itemize}}

%\indent On a final note, although in our experiments the uncertainty calibration metrics are used considering a fixed IoU threshold of 0.5, these metrics can be evaluated across different IoU thresholds, as is occasionally carried out with mAP evaluation.

\bibliographystyle{splncs04}
\bibliography{egbib}
\end{document}

\maketitle

\section*{Additional results}

\begin{figure}[h!]

    \includegraphics[width=1\linewidth]{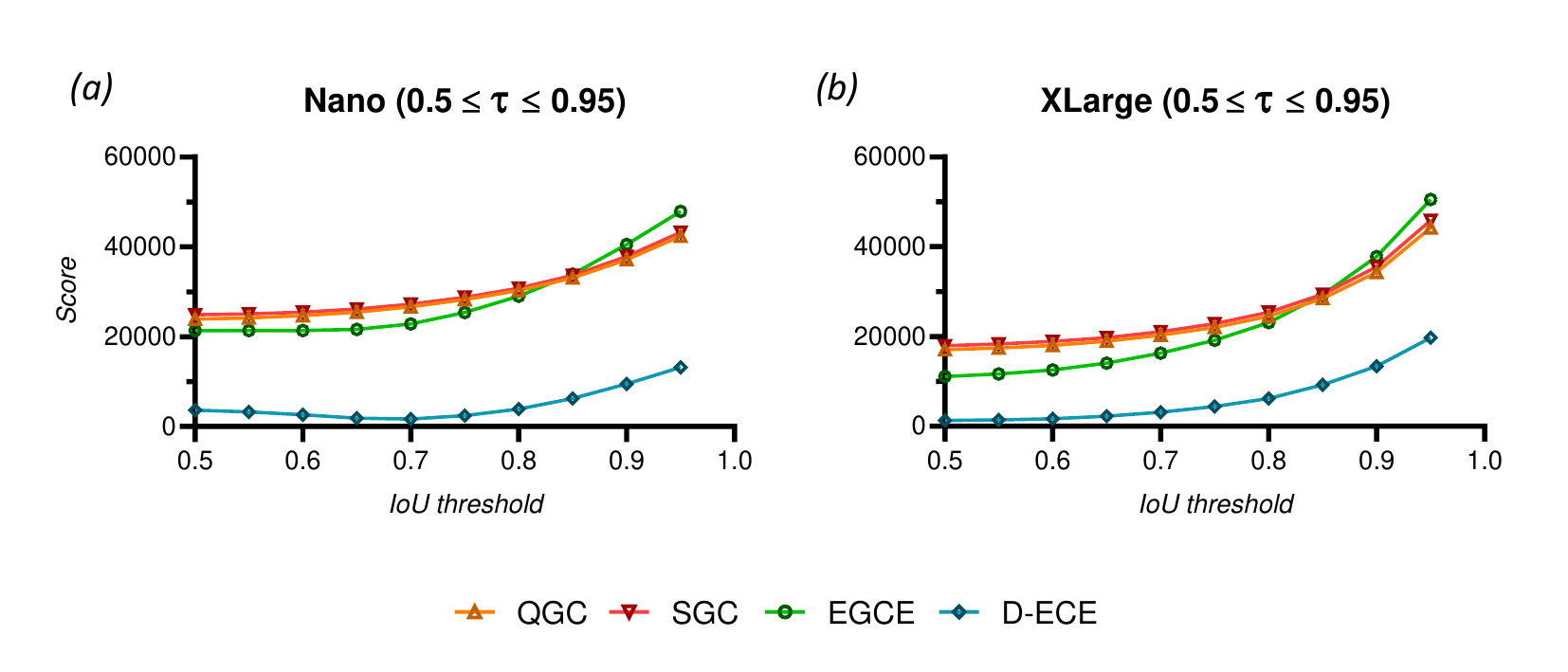}  
    \caption{Evaluating QGC, SGC, EGCE and D-ECE, with increasing values of the IoU threshold $\tau$ (from 0.5 to 0.95, in steps of 0.05). The COCO dataset is used with \textbf{a)} YOLOv5 Nano and \textbf{b)} YOLOv5 Extra Large, for comparison.}
    \label{fig_IoU}
\end{figure}

\noindent In Supp. Figure \ref{fig_IoU}, we contrast the results obtained with the \textit{Nano} (Supp. Figure \ref{fig_IoU}.a) and \textit{Extra Large} (Supp. Figure \ref{fig_IoU}.b) YOLOv5 versions, in the context of uncertainty calibration evaluation with increasing IoU threshold values. This results can shed some light on those discussed in Subsection 5.1 of the main document, specifically on the differences found between the proposed metrics (QGC, SGC, EGCE) and the D-ECE, when we average across different IoU thresholds.\\
\indent It is observable, with any of the metrics, that the aggravation of score in higher thresholds is stronger for the \textit{Extra Large} model than for the \textit{Nano} model; in fact, for every uncertainty calibration metric, there is a threshold above which the \textit{Extra Large} model has worse performance than the \textit{Nano} model. Nonetheless, for the proposed \textit{global} calibration metrics (QGC, SGC, EGCE), this is only true for IoU threshold values above 0.9; contrarily, with the D-ECE, this phenomenon is observable for threshold values starting from 0.65. This justifies what can be observed in Figure 1.b of the main document, where the D-ECE increases as the capacity (and mAP performance) of the model also increases, contrarily to the other uncertainty calibration metrics.
\\

\section*{Distribution-shifts}

The distribution-shifts used in Subsection 5.3 of the main document are created by inducing - in the COCO and PASCAL VOC test sets - \textit{artificial} shifts in distribution, through the injection of \textit{Gaussian noise} and \textit{Gaussian blur}, in 5 different levels of intensity. Specifically, for $i \in \{1,\ldots,5\}$, where $i$ represents a level of shift intensity:
\begin{itemize}
    \item The \textit{Gaussian noise} perturbation has a mean of 0 and variance taken from $U(20i, 100 + 80i)$ for each input, independently for each channel.
    \item The \textit{Gaussian blur} perturbation blurs the input image using a Gaussian filter with a random kernel size within the interval $[-1+2i$,$1+2i]$.
\end{itemize}

\section*{Calibration strategies}

In this Supplementary Section we discuss further the calibration strategies used in Subsection 5.3 of the main document.

\subsection*{Histogram binning}

\textit{Histogram binning} \cite{zadrozny2001obtaining} is a non-parametric method commonly used to calibrate uncertainty estimates of classification models. In \cite{kuppers2020multivariate} this method is adapted to the context of object detection, by substituting the the concept of \textit{average accuracy} per bin, with \textit{precision} per bin, showing positive effects in terms of D-ECE.\\
\indent Specifically, we create a set of bins $\{ B_1, B_2, \ldots, B_M \}$ with boundaries
\begin{align}
    0=a_0<a_1<\ldots<a_M=1
\end{align}
such that, for each confidence value $p$
\begin{align}
p \in B_i, \qquad \text{if } \quad a_{i-1}< p \leq a_i, 
\end{align}
After defining the binning scheme, the \textit{histogram binning} method replaces each prediction $p$ with a new prediction
\begin{align}
prec(B_i), \qquad \text{if } \quad a_{i-1}< p \leq a_i, 
\end{align}
where $prec(B_i)$ is defined as the \textit{precision} calculated with the set of detections that belong to $B_i$. Each \textit{precision} value is calculated in a training set and those values are used at inference, replacing the confidence scores. The bin boundaries are, in our case, defined  as equally spaced intervals, and $M=15$.

\subsection*{Test time augmentation}

TTA refers to the set of techniques that leverage the use of data augmentation ate inference time (test time), \ie, applying different augmentations to the input (often resulting in multiple inputs) before passing it through the model for inference. Intelligently combining multiple outputs, from such augmentations, has shown evidence of improving uncertainty calibration in classification problems \cite{Conde_2022_BMVC}.\\
\indent In the case of this article, TTA is applied with the standard YOLOv5 policy \cite{glenn_jocher_2022_7347926}, by applying horizontal inversions in 3 different scaling version (full, 0.83, 0.67), combining the resulting outputs before \textit{Non Maximum Suppression}.

\section*{Code}

In following outlined files, is the Python code necessary to apply the uncertainty calibration evaluation metrics outlined in the article (the novel metrics QGC, SGC and EGCE, and also the D-ECE). Specifically, the metrics are in the file \textbf{metrics.py}, while the file \textbf{utils.py} has some necessary additional code. To apply the metrics, the input needed is the directory of a folder with all the ground-truth labels (\textit{labels\_folder}) and the directory of a folder with all the predicted bounding boxes (\textit{preds\_folder}); these folders must be in the common YOLOv5 format.\\
\indent Additionally, the code for the version of \textit{histogram binning} used in the article is also made available in the \textbf{histogram\_binning.py} file. For both the training (\textit{HB\_train} function) and prediction (\textit{HB\_predict} function) phases of the method, the folders must be organized in the previously referred form. The output of the \textit{HB\_train} function is used as the second argument of the \textit{HB\_predict} function. The detection bounding boxes with the new confidence scores are saved in a new folder \textit{detections\_hb}, on the current directory.\\

\noindent{\large \textbf{utils.py}}
{\footnotesize
\begin{lstlisting}
import torch
import numpy as np
import os

def text_to_torch(txt_file):
    array = np.loadtxt(txt_file)
    if array.ndim == 2:
        return torch.from_numpy(array)
    if array.ndim == 1:
        return torch.unsqueeze(torch.from_numpy(array),0)
        

def IoU(boxes_preds, boxes_labels):


    box1_x1 = boxes_preds[..., 0:1] - boxes_preds[..., 2:3] / 2
    box1_y1 = boxes_preds[..., 1:2] - boxes_preds[..., 3:4] / 2
    box1_x2 = boxes_preds[..., 0:1] + boxes_preds[..., 2:3] / 2
    box1_y2 = boxes_preds[..., 1:2] + boxes_preds[..., 3:4] / 2
    box2_x1 = boxes_labels[..., 0:1] - boxes_labels[..., 2:3] / 2
    box2_y1 = boxes_labels[..., 1:2] - boxes_labels[..., 3:4] / 2
    box2_x2 = boxes_labels[..., 0:1] + boxes_labels[..., 2:3] / 2
    box2_y2 = boxes_labels[..., 1:2] + boxes_labels[..., 3:4] / 2


    x1 = torch.max(box1_x1, box2_x1)
    y1 = torch.max(box1_y1, box2_y1)
    x2 = torch.min(box1_x2, box2_x2)
    y2 = torch.min(box1_y2, box2_y2)

    
    intersection = (x2 - x1).clamp(0) * (y2 - y1).clamp(0)
    box1_area = abs((box1_x2 - box1_x1) * (box1_y2 - box1_y1))
    box2_area = abs((box2_x2 - box2_x1) * (box2_y2 - box2_y1))

    return intersection / (box1_area + box2_area - intersection + 1e-6)


def fill_blank_preds(labels_folder, preds_folder):
    l1 = os.listdir(labels_folder)
    l2 = os.listdir(preds_folder)
    res = [x for x in l1 if x not in l2]
    for file in res:
        with open(f'{preds_folder}\\{file}', 'w') as f:
            f.write('-1 0.0 0.0 0.0 0.0 0.0')
    

def divide_detections(labels_folder, preds_folder, iou_thresh=0.5):
    fill_blank_preds(labels_folder, preds_folder)
    TP = []
    FP = []
    FN_count = 0
    
    for filename in os.listdir(labels_folder):
        
        TP_file = []
        
        labels_file = os.path.join(labels_folder, filename)
        preds_file = os.path.join(preds_folder, filename)
        labels_tensor = text_to_torch(labels_file)
        preds_tensor = text_to_torch(preds_file)
        
        #TRUE POSITIVES and FALSE NEGATIVES
        for i in range(labels_tensor.size(dim=0)):
            idx = -1
            iou = 0
            for j in range(preds_tensor.size(dim=0)):
                if (labels_tensor[i,0] == preds_tensor[j,0] and 
                    IoU(labels_tensor[i,1:5],preds_tensor[j,1:5]) > iou and 
                    IoU(labels_tensor[i,1:5],preds_tensor[j,1:5]) >= iou_thresh):
                    
                    iou = IoU(labels_tensor[i,1:5],preds_tensor[j,1:5])
                    idx = j
            if idx < 0: #if there is no match, add to False Negatives count
                FN_count += 1
            else: #append to True Positives
                TP_file.append(np.array([preds_tensor[idx,5],iou,idx], dtype=object))
                
        TP.extend(TP_file)

        #FALSE POSITIVES
        for i in range(preds_tensor.size(dim=0)):
            c = -1
            for j in range(len(TP_file)):
                if i == TP_file[j][2]:
                    c = 1
            if c < 0 and preds_tensor[i,0] != -1:
                FP.append(np.array([preds_tensor[i,5]]))
     
    TP = np.array(TP)    
    FP = np.array(FP)
    FN_count = np.array([FN_count])
     
    return TP, FP, FN_count
\end{lstlisting}}
\vspace{3cm}
\noindent{\large \textbf{metrics.py}}

{\footnotesize
\begin{lstlisting}
import numpy as np
import os
from utils import divide_detections

def QGC(labels_folder, preds_folder, iou_thresh=0.5):
    
    TP, FP, FN_count = divide_detections(labels_folder, preds_folder, iou_thresh=iou_thresh)
    
    TP = TP[:,0].astype(float)
    FP = FP.astype(float)
    FN_count = FN_count[0]
    
    tp_score = 0.0
    fp_score = 0.0
    for i in range(len(TP)):
        tp_score += (1-TP[i])**2
    for i in range(len(FP)):
        fp_score += FP[i]**2
    
    QGC_total = tp_score.item() + fp_score.item() + FN_count
    QGC_avg = (tp_score.item() + fp_score.item() + FN_count) / (len(TP) + len(FP) + FN_count)
    
    return QGC_total, QGC_avg



def SGC(labels_folder, preds_folder, iou_thresh=0.5):
    
    TP, FP, FN_count = divide_detections(labels_folder, preds_folder, iou_thresh=iou_thresh)
    
    TP = TP[:,0].astype(float)
    FP = FP.astype(float)
    FN_count = FN_count[0]
    
    tp_score = 0
    fp_score = 0
    for i in range(len(TP)):
        tp_score += TP[i] / (TP[i]**2 + (1-TP[i])**2)**(1/2)
    for i in range(len(FP)):
        fp_score += (1-FP[i]) / (FP[i]**2 + (1-FP[i])**2)**(1/2)
      
    SGC_total = len(TP) + len(FP) + FN_count - tp_score.item() - fp_score.item()
    SGC_avg = ((len(TP) + len(FP) + FN_count - tp_score.item() - fp_score.item()) / 
               (len(TP) + len(FP) + FN_count))
    
    return SGC_total, SGC_avg




def EGCE(labels_folder, preds_folder, num_bins=15, iou_thresh=0.5, conf_thresh=0.1):
         
    TP, FP, FN_count = divide_detections(labels_folder, preds_folder, iou_thresh=iou_thresh)
    
    TP = TP[:,0].astype(float)
    FP = FP.astype(float)
    FN_count = FN_count[0]
    
    TP_new = TP[TP>=conf_thresh]
    FP = FP[FP>=conf_thresh]
    
    FN_count = FN_count + (len(TP)-len(TP_new))
    
    TP=TP_new
    
    bins = np.linspace(0.0, 1.0, num_bins+1)#[1:]
    bins[0] = bins[0]-0.001

    TP_binned = np.digitize(TP, bins, right=True)-1
    FP_binned = np.digitize(FP, bins, right=True)-1
    
    EGCE = 0
    
    bin_precs = np.zeros(num_bins)
    bin_confs = np.zeros(num_bins)
    bin_sizes = np.zeros(num_bins)
    
    for i in range(num_bins):
        if i == num_bins-1:
            bin_sizes[i] = (len(TP_binned[TP_binned == i]) + len(FP_binned[FP_binned == i]) + 
                            FN_count)
        else:
            bin_sizes[i] = len(TP_binned[TP_binned == i]) + len(FP_binned[FP_binned == i])
        
        if bin_sizes[i] > 0:
          
          if i == num_bins-1:
              bin_precs[i] = len(TP_binned[TP_binned == i]) / bin_sizes[i]
              bin_confs[i] = (( (TP[TP_binned==i]).sum() + (FP[FP_binned==i]).sum() 
                             + FN_count) / bin_sizes[i])
          else:
              bin_precs[i] = len(TP_binned[TP_binned == i]) / bin_sizes[i]
              bin_confs[i] = (( (TP[TP_binned==i]).sum() + (FP[FP_binned==i]).sum() ) 
                              / bin_sizes[i])
          
    for i in range(num_bins):
      abs_conf_dif = abs(bin_precs[i] - bin_confs[i])
      EGCE += (bin_sizes[i] ) * abs_conf_dif
      
    
    EGCE_avg = EGCE/sum(bin_sizes)
      
    return EGCE, EGCE_avg
    

    
def DECE(labels_folder, preds_folder, num_bins=15, iou_thresh=0.5, conf_thresh=0.1):
         
    TP, FP, _ = divide_detections(labels_folder, preds_folder, iou_thresh=iou_thresh)
    
    TP = TP[:,0].astype(float)
    FP = FP.astype(float)
    
    TP = TP[TP>=conf_thresh]
    FP = FP[FP>=conf_thresh]
    
    bins = np.linspace(0.0, 1.0, num_bins+1)#[1:]
    bins[0] = bins[0]-0.001

    TP_binned = np.digitize(TP, bins, right=True)-1
    FP_binned = np.digitize(FP, bins, right=True)-1
    
    DECE = 0
    #DECE = 0
    
    bin_precs = np.zeros(num_bins)
    bin_confs = np.zeros(num_bins)
    bin_sizes = np.zeros(num_bins)
    
    for i in range(num_bins):
        bin_sizes[i] = len(TP_binned[TP_binned == i]) + len(FP_binned[FP_binned == i])
        if bin_sizes[i] > 0:
            bin_precs[i] = len(TP_binned[TP_binned == i]) / bin_sizes[i]
            bin_confs[i] = (( (TP[TP_binned==i]).sum() + (FP[FP_binned==i]).sum() ) / 
                            bin_sizes[i])
          
    for i in range(num_bins):
      abs_conf_dif = abs(bin_precs[i] - bin_confs[i])
      DECE += (bin_sizes[i] ) * abs_conf_dif
      
    
    DECE_avg = DECE/sum(bin_sizes)
      
    return DECE, DECE_avg
\end{lstlisting}}
\vspace{2cm}
\noindent{\large \textbf{histogram\_binning.py}}

{\footnotesize
\begin{lstlisting}

import numpy as np
from utils import divide_detections, text_to_torch
import os 
from os.path import join


def HB_train(labels_folder, preds_folder, num_bins=15, iou_thresh=0.5):
         
    TP, FP, _ = divide_detections(labels_folder, preds_folder, iou_thresh=iou_thresh)
    
    TP = TP[:,0].astype(float)
    FP = FP.astype(float)
    
    bins = np.linspace(0.0, 1.0, num_bins+1)#[1:]
    bins[0] = bins[0]-0.001

    TP_binned = np.digitize(TP, bins, right=True)-1
    FP_binned = np.digitize(FP, bins, right=True)-1
    
    bin_precs = np.zeros(num_bins)
    bin_sizes = np.zeros(num_bins)
    
    for i in range(num_bins):
        bin_sizes[i] = len(TP_binned[TP_binned == i]) + len(FP_binned[FP_binned == i])
        if bin_sizes[i] > 0:
            bin_precs[i] = len(TP_binned[TP_binned == i]) / bin_sizes[i]
            
    return bin_precs




def HB_predict(preds_folder, bin_precs, num_bins=15):
    
    bins = np.linspace(0.0, 1.0, num_bins+1)#[1:]
    bins[0] = bins[0]-0.001
    
    out_folder = "detections_hb"
    
    if not os.path.exists(out_folder):
        os.makedirs(out_folder)
    
    for filename in os.listdir(preds_folder):
        
        preds_file = os.path.join(preds_folder, filename)
        pred_list = text_to_torch(preds_file).tolist()
        
        for n in range(len(pred_list)):  
            for i in range(num_bins):
                
                if bins[i] < pred_list[n][5] <= bins[i+1]:
                    pred_list[n][5] = bin_precs[i]
                    break              
                
        with open(join(out_folder,filename), 'w') as fp:
            for item in pred_list:
                for i in item:
                    fp.write("%s " %i)
                fp.write("\n")

\end{lstlisting}}

\bibliographystyle{splncs04}
\bibliography{egbib}